\documentclass[12pt,draftcls,peerreview,onecolumn]{IEEEtran}
\usepackage{booktabs}
\usepackage{amsmath,amssymb,amsfonts}
\usepackage{algorithm}
\usepackage{algpseudocode}
\usepackage{graphicx}
\usepackage{xcolor}
\begin{document}
%
\title{Smart Scheduling based on Deep Reinforcement Learning for Cellular Networks}
\author{\IEEEauthorblockN{Jian ~Wang\IEEEauthorrefmark{1}, Chen ~Xu\IEEEauthorrefmark{1}, Rong ~Li\IEEEauthorrefmark{1}, Yiqun ~Ge\IEEEauthorrefmark{2}, Jun ~Wang\IEEEauthorrefmark{1}}\\
\IEEEauthorblockA{\IEEEauthorrefmark{1}Hangzhou Research Center, Huawei Technologies, Hangzhou, China}\\
\IEEEauthorblockA{\IEEEauthorrefmark{2}Ottawa Research Center, Huawei Technologies, Ottawa, Canada}\\
Emails: \{wangjian23, xuchen14, lirongone.li, yiqun.ge, justin.wangjun\}@huawei.com
}

\maketitle
\begin{abstract}
To improve the system performance towards the Shannon limit, advanced radio resource management mechanisms play a fundamental role. In particular, scheduling should receive much attention, because it allocates radio resources among different users in terms of their channel conditions and QoS requirements. The difficulties of scheduling algorithms are the tradeoffs need to be made among multiple objectives, such as throughput, fairness and packet drop rate. We propose a smart scheduling scheme based on deep reinforcement learning (DRL). We not only verify the performance gain achieved, but also provide implementation-friend designs, i.e., a scalable neural network design for the agent and a virtual environment training framework. With the scalable neural network design, the DRL agent can easily handle the cases when the number of active users is time-varying without the need to redesign and retrain the DRL agent. Training the DRL agent in a virtual environment offline first and using it as the initial version in the practical usage helps to prevent the system from suffering from performance and robustness degradation due to the time-consuming training. Through both simulations and field tests, we show that the DRL-based smart scheduling outperforms the conventional scheduling method and can be adopted in practical systems.
\end{abstract}
\begin{IEEEkeywords}
artificial intelligence, cellular networks, deep reinforcement learning, smart scheduling
\end{IEEEkeywords}

\IEEEpeerreviewmaketitle

\section{Introduction}
\label{sec1}
The wireless communication industry has been keeping a fast growing and updating speed for several decades. About every ten years, new generations of mobile communication system were standardized with lots of new features and supported scenarios. Thanks to the evolution of wireless communications technologies, we are now enjoying diverse services and applications conveniently. It is well known that the fifth generation (5G) mobile communications system supports three major categories of services, i.e., enhanced mobile broadband (eMBB), ultra-reliable and low-latency communications (uRLLC) and massive machine-type communications (mMTC). Meanwhile, new applications and scenarios have never stopped coming up, which sets up new requirements including even higher throughput, more connected devices, faster access with lower latency and higher efficiency for wireless communication systems. With all these requirements in mind, designing a new generation of mobile communications system becomes a quite challenging work. The overall improvement calls for progress not only in the link level such as new signal processing schemes in the physical layer, but also in the system level such as scheduling and coordination methods in upper layers.

Scheduling is essentially a decision making task, usually modeled and solved through a utility maximization problem. The formulation of the problem is based on the mathematical representation of the whole communication system, which inevitably results into some approximations and assumptions. In the wireless scenario, many factors have impacts on the system modelling, such as user equipment (UE) mobility, traffic model, channel variance, etc. These factors make the wireless communication network a system with high dynamics, hence it is almost impossible for the utility maximization based methods to be optimal across all ranges of scenarios in practice.

Promoted by the advanced algorithm, powerful computing capability and rich-content data sets, artificial intelligence (AI) \cite{russell2016artificial} has been as the driving force of a new wave of technical revolution. As a specific sub-set of AI technologies, reinforcement learning (RL) provides a novel way of solving decision making problems. Instead of an explicit mathematical model, decision optimization in a dynamic system can be modeled into a Markov Decision Process (MDP) whose states and actions are led by a reward to be defined. A RL agent approaches to an optimal solution of a MDP by learning from its interactivity with its true physical environment. It firstly gathers the state information of the physical environment, yields the best policy to maximize the reward, and makes a series of actions based on the policy. The agent can adjust its policy continuously until the best policy is approached. The usage of conventional RL algorithms such as Q-learning is usually limited by the dimension of the problem. To handle practical problems with large state and action spaces, DNNs are recently introduced to be used as RL agents, which brings the technique named deep reinforcement learning (DRL). In this paper, to address the scheduling problem in cellular networks, we propose a deep reinforcement learning (DRL) based method. To be specific, the scheduling problem is modelled into an MDP and solved through DRL. The state of the MDP includes the channel-related parameters indicating the current wireless channel quality and the buffer-related ones reflecting the Quality-of-Service (QoS) requirement. The DRL agent is trained to make decisions to maximize the reward, a joint consideration of throughput, fairness and packet loss performance. As a result of scheduling each involved UE, the action issued by the DRL agent are realized by the scheduling signalling from BS to UEs.

To embed the proposed DRL based scheduling scheme into practical systems, we have to carefully address several issues. Firstly, compared with the conventional scheduling schemes, the proposed one should have superior performance. Otherwise, it would be unreasonable to conduct algorithm replacement. Secondly, the sporadic arrival and departure of UEs for a target cell keep resizing the scheduling problem. Accordingly, the DRL agent should be scalable enough to avoid retraining new agents whenever the scheduling problem is resized. Thirdly, the training of the DRL agent is time-consuming, in which exploration of potential actions may results in really bad performance. In facing the performance and robustness degradation due to the exploration, AI-based methods should be accelerated. Finally, the action decisions should be conducted in time, which have the inference completed within the scheduling period.

With the practical considerations mentioned above and some preliminary results published in \cite{wang2019deep} and \cite{xu2019buffer}, we do contributions in this paper as follows:
\begin{itemize}
  \item We propose a DRL based smart scheduling scheme for cellular networks. Both the wireless channel quality and the packet buffer condition are included into the state of the system. Throughput, fairness and packet loss performance are jointly considered and optimized. With improvements to facilitate the training and inference of the DRL agent, the proposed scheme outperforms baseline in both simulation and field test. Two gene-aided scheduling schemes with global and future information available are elaborated. Through comparisons between the proposed scheme and the two gene-aided ones, the performance gain space of the scheduling problem is verified.
  \item We propose a scalable DRL agent, which exploits the same fixed-size DNN structure in different deployments with different numbers of UEs. Thanks to the scalability, we neither limits the maximum number of UEs nor re-train the agent when the scheduling problem is resized.
  \item We propose a virtual environment training framework. The DRL agent can be firstly trained in a background virtual environment that is built upon information collected from the practical system. After converging in the virtual environment, the pre-trained DNN parameters can be deployed as the initial version of the DRL agent in practice. During the background training, some conventional algorithms can be used in the practical system to prevent the system from adopting a not-well-trained agent and experiencing disastrous performance.
  \item We build up both simulation platform and field test prototype, and verify that the performance gain of the proposed smart scheduling scheme is stable in both simulation and field test. More importantly, through the field test, we prove that the proposed DRL agent can make scheduling decisions in time within a scheduling period of 10ms.
\end{itemize}

The rest of this paper is organized as follows. We review the related work in Section II. The system model and problem formulation are described in Section III. The exploration methods of the performance gain space are shown in Section IV. In Section V, we propose the DRL based scheduling scheme. Then, the evaluations of the proposed scheme are done in both simulation and test field. The results and discussions are provided in Section VI. Finally, we conclude the paper in Section VII with some future work directions provided.

\section{Preliminaries and Related Work}
\label{sec2}
A scheduler, usually equipped in the BS, is the brain of a cellular network, because it makes the important decision on how radio resources are allocated among users. Typical scheduling schemes can be classified into tree types. The simplest ones stemmed from wired networks, without considering any channel conditions. Round-Robin (RR) is the most famous algorithm in this category \cite{rasmussen2008round}. Radio resources are allocated among UEs with equal probability, no matter how good or bad the channels are. This type cannot perform adaptive adjustment according to channel changing, a typical phenomenon in wireless communications. Thanks to the design of channel quality indicator (CQI) feedback in cellular networks, the scheduler is aware of the channel conditions of the UEs. Then, the second type was designed to take use of this channel quality information. For example, maximum Carrier-to-Interference (Max C/I) scheme aims at maximizing the overall throughput by allocating the resources to UEs with the best channel conditions. This strategy sacrifices the fairness among UEs, because those UEs with poor channel conditions may always be discriminated. The third type takes both channel conditions and QoS requirements into account. There always exists UEs with different QoS requirements in the practical systems. QoS parameters that represents the QoS requirements of a UE and the CQI feedbacks from that UE would be input into the scheduling algorithm in form of data rates and delays as part of the scheduling metrics. A good survey can be found in \cite{capozzi2013downlink}, where it is pointed out that an optimal tradeoff between throughput and fairness is usually pursued while designing scheduling schemes.

Proportional Fair (PF), belonging to the second category aforementioned, is among the most widely used scheduling algorithms. Compared with Max C/I scheme, PF provides a tradeoff between system overall throughput and fairness among UEs. F. Kelly provides a general principle about how to achieve proportional fairness by formulating the scheduling problem into an utility maximization problem
\begin{equation}\label{eq:ump}
\begin{split}
  \max \quad &\sum\limits_{n \in \cal{N}} {U_n \left( {x_n } \right)} \\
  {\rm{s}}{\rm{.t}}{\rm{.}}\quad & \;\;x_n \;{\rm{is}}\;{\rm{feasible}}
\end{split}
\end{equation}
where, $x_n, n \in \cal{N}$ is the data rate of the $n$-th UE. The feasible condition is that $x_n \geq 0$ and sum of $x_n$ is no larger than the system capacity $C$. When the utility function $U_n$ is a logarithm function, the solution of problem (\ref{eq:ump}) has the unique vector of rates, named as proportionally fair \cite{kelly1997charging}. Therefore, the PF scheduling algorithm is optimal in term of maximizing the sum of logarithmic rate.

Later, \cite{tse2001multiuser} adopts it into the scenario of wireless networks with a single carrier. The only carrier (or say channel) should be allocated to the UE with the largest metric of $I_n /T_n $, where $I_n$ is the instantaneous throughput estimated from the updated channel condition and $T_n$ is the average throughput within a past time window for the $n$-th UE, respectively. This metric indicates that the UEs with higher instantaneous throughputs (i.e., better channel conditions) should be given higher priorities to access the channel resource, thereby improving the overall throughput of the system. Proportionally, the UEs with smaller historic throughputs should be given more chance to access the channel, thereby guaranteeing some degree of fairness. To summarize, for a single-carrier system, to achieve proportional fairness, UE should be chosen according to
\begin{equation}\label{eq:pf}
i = \mathop {\arg \max }\limits_{n \in \cal{N}} {{I_n } \over {T_n }}
\end{equation}
where, the average throughput is updated according to
\begin{equation}\label{eq:ave_rate}
T_{n}\left ( t \right ) = \frac{W-1}{W}T_{n}\left ( t-1 \right ) + \frac{1}{W}I_{n}\left ( t \right )
\end{equation}
with $W$ as the window size for averaging.
For multi-carrier systems, \cite{kim2005proportional} provides the method to achieve proportional fairness, and a greedy way is elaborated in \cite{sun2006reduced}.

Starting from the PF scheduling scheme, more and more scheduling schemes are designed with the considerations of satisfying the QoS requirements. \cite{ramli2009performance} proposed a modified version of largest weighted delay first (M-LWDF) scheme that tries to ensure a good balance among throughput, fairness and accumulated delay. A buffer-aware and traffic-dependent scheduling scheme is proposed in \cite{huang2007buffer}, where the original PF metric calculation expression is modified to involve packet arrival rate, remaining buffer space and head-of-line (HoL) packet size in an intuitive way. Besides spectral efficiency, energy efficiency is also taken into account while designing scheduling scheme in \cite{yang2015survey}. It is pointed out that PF is feasible for stationary channels, however mobility of UEs and slow-fading of channels bring non-stability. \cite{margolies2014exploiting} provides scheduling schemes with prediction capability to fix this issue. \cite{tsai2008state} considers the scenario where UEs may newly join the network or resume from the idle state. The average rate calculation, originally shown as Eq. (\ref{eq:ave_rate}), is replaced for the newly coming UEs.

The above scheduling schemes are all deterministic ones, requiring explicit models or utility functions and appear as deterministic expressions. However, they suffer from a low flexibility when adopted in mobile communication systems with high dynamics. An alternative way is to model the scheduling problem into an MDP, whose scheduling decisions are made based on the observations of the system states. The state space of the MDP can be designed to reflect the dynamics of the system. Although MDP can be solved by Dynamic Programming (DP) \cite{bertsekas1995dynamic}, it needs the transition probabilities of states as a priori, which is generally hard to acquire in the scheduling problem. To bypass this issue, DRL can be used as the solver of the scheduling MDP problem. With the help of DNNs, DRL can handle MDP problem with large state and action spaces.

DRL has already been used in wireless communication systems, and a comprehensive survey can be found in \cite{luong2019applications}. The most relevant work for this paper are those about using DRL to do scheduling. In \cite{atallah2017deep}, the DRL-based scheduler is introduced to extend the lifetime of the battery-powered Road-Side Units (RSUs) while promoting a safe environment that meets acceptable QoS levels in a Vehicle-to-Infrastructure (V2I) scenario. Small-cell BS (SBS) selection and content caching tasks are done simultaneously by a DRL-based scheduler in \cite{wei2018joint}. In \cite{zhu2018new}, a DRL-based scheduler is designed to coordinate packet transmission from different buffers through multiple channels in a cognitive IoT network. Uplink scheduling is studied in \cite{chu2018reinforcement}, where throughput maximization is achieved by properly choosing $K$ from $N$ UEs. Two exponential parameters are introduced to modify the numerator and denominator of Eq. (\ref{eq:pf}), and DRL is used to tune these parameters to achieve different levels of tradeoff between throughput and fairness in \cite{comcsa2014adaptive}. \cite{chinchali2018cellular} proposes to use a DRL-based scheduler to optimize the Internet of Things (IoT) traffic without impacting conventional real-time applications such as voice-calling and video. Real field data are gathered to train the agent, which is then evaluated in field tests.

Most existing works focus on coordination between multiple types of data traffic (IoT and eMBB) or multiple levels of networks (heterogeneous networks). The proposed schemes are trained and evaluated on a simulation environments with quite loose assumptions. For instance, data rate is derived directly from Shannon's equation, without considering adaptive modulation and coding (AMC) and outer loop link adaption (OLLA). Moreover, existing DRL algorithms are used directly without much specific design for the DNNs used as agents in these references. In contrast to them, this paper proposed a smart scheduling scheme for cellular networks with a specific scalable design of DNNs for the DRL agent. The training and evaluation are done both on a system simulation platform and through a field test prototype.

\section{System Model}
\label{sec3}

\begin{figure}[!t]
	\centering
	\includegraphics[width=8cm]{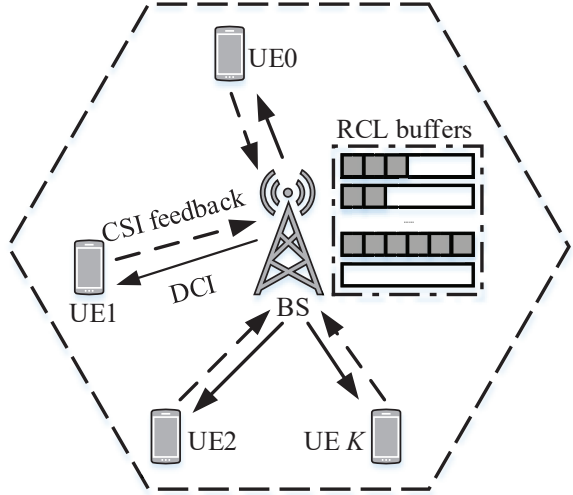}
	\caption{System model.}
	\label{fig:systemmodel}
\end{figure}

In this paper, we consider the downlink scheduling in a single cell cellular network as shown in Fig.~\ref{fig:systemmodel}. The BS equipped with a scheduler allocates radio resources among $K$ UEs. The available radio resources are divided into $B$ resource block groups (RBGs), each of which can be occupied by only one UE during each transmission time interval (TTI). The BS maintains one Radio Link Control (RLC) layer buffer for each UE so that the packets for this UE from upper layers are stored for transmission or retransmission. As the full buffer traffic mode case is well discussed in our previous paper \cite{wang2019deep}, we mainly consider the more practical non-full buffer traffic modes in this paper. In the non-full buffer mode, packets follows a Poisson process with arrival rate $\lambda$. The packets may be eliminated due to either buffer overflow or expiration in the queue. UEs feed their channel state informations (CSIs) back to the BS to help the BS to make scheduling decisions. The scheduling results are sent to the UEs through downlink control information (DCI). Both the feedback and scheduling are on the TTI basis. A UE becomes inactive when its RLC buffer is empty. This inactive UE will not participate into the next round of scheduling until new packets arrive at its RLC buffer, after then it becomes active again.

As mentioned in \cite{capozzi2013downlink}, the optimal tradeoff between throughput and fairness is always the aim of scheduling scheme design. Meanwhile, for non-full buffer mode, packet loss due to buffer overflow and expiration is also a key performance metric for practical finite buffer systems. Hence, in this paper, we consider throughput (THP), fairness (indicated by Jain's fairness index \cite{jain1984quantitative} (JFI)) and packet drop rate (PDR) as three key performance indicators (KPIs), which can be expressed as
\begin{equation}
\label{eq:kpi}
\begin{split}
&\mathrm{THP} = \sum \limits_t \sum \limits_{k \in \cal{K}} \sum \limits_{b \in \cal{B}} d_{k,b}(t)r_{k, b}(t)\\
&\mathrm{JFI} = \dfrac{\left[\sum \limits_t \sum \limits_{k \in \cal{K}} \sum \limits_{b \in \cal{B}} d_{k,b}(t)r_{k, b}(t)\right]^2}{K \sum \limits_{k \in \cal{K}} \left[\sum \limits_t  \sum \limits_{b \in \cal{B}} d_{k,b}(t)r_{k, b}(t)\right]^2}\\
&\mathrm{PDR}=\dfrac{\sum \limits_t \sum \limits_{k \in \cal{K}} (a_k(t) - s_k(t))}{\sum \limits_t \sum \limits_{k \in \cal{K}} a_k(t)}
\end{split}
\end{equation}
where $\cal K$ and $\cal B$ are the set of UEs and RBGs, respectively. $r_{k, b}(t)$ denotes the achievable rate of the $k$-th UE at the $b$-th RBG at the $t$-th TTI, $d_{k,b}(t)\in\{0,1\}$ is the scheduler decision about whether the $b$-th RBG is allocated to the $k$-th UE. $a_k(t)$ and $s_k(t)$ are the numbers of arrived packets and transmitted packets for the $k$-th UE, respectively. All the three KPIs are statistically computed over a sufficiently long period, e.g., several hundreds of TTIs, so that the scheduler algorithm should pay more attention to the long-term reward than the short-term one. We define this long period as a scheduling duration, within which the proposed scheme and the baseline are compared.

\section{Exploration of Pareto Frontier}
\label{sec4}
Simultaneously considering multiple KPIs turns scheduling into a multi-objective optimization problem. Theoretically, Pareto optimization can be used to find all the nondominant tradeoffs among optimization objectives. All the tradeoff solutions consist into a Pareto frontier, each point of which is an optimal tradeoff subjected to a typical circumstance.

For the full buffer traffic mode, the aim of solving scheduling problem is to find an optimal tradeoff between throughput and fairness, while packet loss in the RLC buffer is not considered. Two extreme cases on the Pareto frontier in this case are the Max C/I scheduling and Max-Min scheduling. Max C/I scheduling guarantees the highest throughput performance without considering fairness, whereas Max-Min scheduling provides the highest level of fairness with lowest throughput by trying to maximizing the throughput of the UE with lowest historic throughput. Between these two solutions, PF scheduling provides a good tradeoff between throughput and fairness. It has been proved that, for one carrier (or channel, or RBG) systems with full buffer traffic mode, PF scheduling is also on the Pareto frontier, which justify itself as an optimal tradeoff \cite{kelly1997charging}.

Taking packet loss into consideration with the non-full buffer traffic mode, the Pareto frontier contains the optimal tradeoffs among throughput, fairness and packet drop rate. These three KPIs are all long-term statistics as shown in Eq. (\ref{eq:kpi}). Although THP is linear and additive across the scheduling periods (TTIs), it is uneasy to tackle with JFI and PDR since the scheduling decision for the current TTI may have impacts on the following ones. A greedy scheduling scheme that chooses the decision with the highest JFI or lowest PDR in the current TTI, may not achieve the best long-term JFI or PDR performance in the end. In this situation, it is difficult to find out the complete Pareto frontier for the scheduling problem. For a scheduling duration of $N$ TTIs in a system with only one RBG, the scheduling decisions for each TTI constitute a sequence with the length of $N$. Suppose that the complete channel conditions and RLC buffer conditions of the $K$ UEs in these $N$ TTIs be known to the scheduler \emph{a priori}, the searching space contain totally $K^N$ candidates, i.e., $N$ scheduling decisions, each of which has $K$ options. It is obvious that to find the optimal solution by an exhaustive search in such a huge space is impossible for large $K$ and $N$.

In this section, we propose two heuristic gene-aided algorithms to explore the performance gain space of scheduling problem with non-full buffer traffic mode, i.e., genetic algorithm (GA) and Pareto list algorithm (PLA). By the word ``gene-aided'', we assume that the channel qualities and RLC buffer conditions for the $N$ TTIs be known \emph{a priori} at the beginning of the scheduling duration. They can help to explore the Pareto frontier in this multi-objective scheduling problem. The two algorithms are elaborated in following subsections with an assumption that there be only one RBG. The algorithms can be simply extended to multi-RBG cases by scheduling the UEs in a RBG-by-RBG way.

\subsection{Genetic Algorithm}
\label{sec4.1}
The $N$ successive scheduling decisions consist into a scheduling sequence, while the optimal one can be searched through Genetic algorithm (GA). In GA, for every generation, $M$ scheduling sequences (population) with the length of $N$ are obtained through selection, crossover and mutation. The sequences are evaluated since all the channel and RLC buffer conditions are known \emph{a priori}. The good ones are selected for the next generation. The main concern during GA process is to obtain multiple scheduling sequences with good performance and high diversity. In this paper, we adopt the nondominated sorting genetic algorithm II (NSGA-II) as shown in Algorithm \ref{alg:NSGA} \cite{deb2002fast}.

\begin{algorithm}[!htbp]
	\caption{NSGA-II algorithm}\label{alg:NSGA}
	\begin{algorithmic}
		\State Set the population size $M$ and the number of generations $G$
        \State Initialize the parent population $\mathcal{P}_1$ and offspring population $\mathcal{Q}_1 = \emptyset$
		\For {$t = 1$ to $G$}
		\State $\mathcal{R}_t = \mathcal{P}_t \cup \mathcal{Q}_t$
		\State $\mathcal{F} =$ \Call{Fast\_Nondominated\_Sorting\_Procedure}{$\mathcal{R}_t$}
		\State $\mathcal{P}_{t+1} = \emptyset$
        \While {$|\mathcal{P}_{t+1}| + |\mathcal{F}_i| \leq M$}
		\State \Call{Crowing\_Distance\_Assignment}{$\mathcal{F}_i$}
		\State $\mathcal{P}_{t+1} = \mathcal{P}_{t+1} \cup \mathcal{F}_i$
        \State $i = i+1$
        \EndWhile
        \State sort($\mathcal{F}_i, \prec_n$)
        \State $\mathcal{P}_{t+1} = \mathcal{P}_{t+1} \cup \mathcal{F}_i\left[1:\left(M - |\mathcal{P}_{t+1}| \right) \right]$
        \State $\mathcal{Q}_{t+1} =$ \Call{Make\_New\_Pop}{$\mathcal{P}_{t+1}$}
        \State $t = t+1$
		\EndFor
	\end{algorithmic}
\end{algorithm}

According to Algorithm \ref{alg:NSGA}, two key operations are adopted. Firstly, the population is sorted into different nondomination levels through the function Fast\_Nondominated\_Sorting\_Procedure($\mathcal{R}_t$). The definition of domination has considered all three objectives. For instance, we say solution $x$ dominates $y$ if the THP of $x$ is higher than that of $y$ with JFI and PDR of $x$ no worse than those of $y$. Solutions with higher nondomination level are chosen with high priority so that the elitism is preserved. Secondly, the crowing distance for each individual in the same nondomination level is assigned through the function Crowing\_Distance\_Assignment($\mathcal{F}_i$). Within the same nondomincation level, solutions with higher crowing distance are chosen first, which in turn guides the selection toward a uniformly spread-out on the Pareto frontier. In the considered scheduling problem, the length of chromosome is $N$, and the variable of each gene lies in $\left\{1,2,\ldots,K\right\}$. The crossover and mutation operations are conducted in the function Make\_New\_Pop($\mathcal{P}_{t+1}$), which introduces possibilities of generating new and better scheduling decision sequences. The details of the functions and crowding-comparison operator $\left(\prec_n\right)$ can be found in \cite{deb2002fast}.

\subsection{Pareto List Algorithm}
\label{sec4.2}
In contrast to GA that outputs the complete scheduling decision sequence in one shot, Pareto list algorithm (PLA) solves the scheduling problem in a tree-like searching way as shown in Algorithm \ref{alg:PLA}.

\begin{algorithm}[!htbp]
	\caption{Pareto list algorithm}\label{alg:PLA}
	\begin{algorithmic}
		\State Set the maximum list number $L_{\max}$
        \State Initialize the initial list size $L_0=1$ and path $\mathcal{O}_{L_0}^0 =\emptyset$
        \For {$t=1$ to $N$}
        \State $L_t=0$
        \For {$l=1$ to $L_{t-1}$}
        \For {$k=1$ to $K$}
        \If {$r_k(t) > 0$}
        \State $L_t = L_t +1$
        \State $\mathcal{\widehat{O}}_{L_t}^t = [\mathcal{O}_{l}^{t-1}, k]$
        \EndIf
        \EndFor
        \EndFor
        \State Calculate $\mathrm{THP}_{L_t}^t$, $\mathrm{JFI}_{L_t}^t$ and $\mathrm{PDR}_{L_t}^t$ for all paths in $\widehat{\mathcal{O}}_{L_t}^t$ using Eq. (\ref{eq:kpi})
        \If {$L_t > L_{\max}$}
        \State Do nondominated and crowing distance sorting in $\widehat{\mathcal{O}}_{L_t}^t$ according to NSGA-II
        \State Prune the path with 0 crowing distance
        \State Preserve the first $L_{\max}$ paths in $\widehat{\mathcal{O}}_{L_t}^t$ to form $\mathcal{O}_{L_t}^t$
        \State $L_t = L_{\max}$
        \EndIf
        \EndFor
        \State Select the best path from $\mathcal{O}_{L_N}^N$
	\end{algorithmic}
\end{algorithm}

The Pareto list algorithm executes path expanding, sorting and pruning TTI by TTI, and constraining the complexity to a fixed level, i.e., the maximum number of the list $L_{\max}$. At each TTI, a path is expanded according to the number of active UEs. Note that different paths may lead to different active UEs and be expanded in different ways, because paths affect the states, such as UE buffers. After several TTIs, the number of survival paths may exceed $L_{\max}$, then path sorting and pruning are done. Unlike the conventional list-based algorithm in which the path metric is a scalar, the path sorting and pruning in the scheduling issue should consider multiple objectives. Again, the nondominated and crowing distance sorting method introduced by NSGA II can be used. Moreover, in the scheduling problem, a large number of paths may result in a same state, implying the exact same impact on the following TTIs. In this case, only one path from those resulting in the same state is kept alive before doing the nondominated and crowing distance sorting. The definition of domination used in PLA is the same to that in GA, i.e., the paths with higher THP and no less JFI/PDR are preferred. After $N$ expansions, a final scheduling sequence is selected from the set of $L_{\max}$ survival paths.

\section{Smart Scheduling based on Deep Reinforcement Learning}
\label{sec5}
\begin{figure}[!t]
	\centering
	\includegraphics[width=8cm]{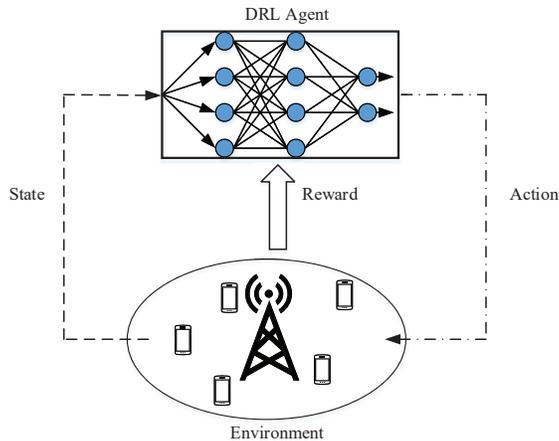}
	\caption{DRL framework for scheduling problem.}
	\label{fig:DRL}
\end{figure}

In DRL algorithms, the DNN-realized agent is trained through interactions with the environment. For the scheduling problem in this paper, the framework of DRL based scheme is shown in Fig.~\ref{fig:DRL}, where the environment is a cellular network. The DRL agent observes the state $s$ of the environment, and makes an action decision $a$ based on its policy $\pi(a|s)$ accordingly. The action is issued into the environment, after which the state transits into the next state and a reward $r$ is obtained. Based on the reward together with the state and action, the DRL agent adjusts its decision policy in order to get higher reward expectation in future. Although the idea of using DRL for scheduling comes easily, there are some challenges in the real design:
\begin{itemize}
  \item When the estimated rate that reflects channel conditions is regarded as continuous value, the state space increases substantially big. It it quite challenging to train a DRL agent in such a huge state space.
  \item When the trade-offs among three long-term KPIs (THO, JFI, and PDF) is considered, it is not straightforward to design a reward function.
  \item Stochastic arrivals and departures of the UEs would result into temporal changes in the state and action space. 
\end{itemize}

In this section, beginning with MDP modelling, we elaborate the smart scheduling scheme with a scalable, easy-training and fast-convergence design.

\subsection{Markov Decision Process}
\label{sec5.1}
A MDP is typically defined by a tuple $\left ({\cal{S}}, {\cal{A}}, P, r \right)$, where $\cal{S}$ is the set of states, $\cal{A}$ is the set of actions, $P(s'|s, a)$ is the transition probability from state $s$ to $s'$ due to action $a$, and $r$ is the immediate reward when transition happens. Although it is hard to obtain the transition probability in this scheduling problem, fortunately the DRL-based scheme doesn't need it. The state, action and reward can be defined as follows.

\textbf{State}: The input state contains observations for each UE. The estimated rate, averaged rate, the spare space in the RLC buffer and the waiting time of the HoL packet are concatenated to form the observation of each UE. Per this definition, it is obvious that the scheduling problem has a continuous state space.

\textbf{Action}: For the single RBG case, the action is a $K$-length one-hot code, indicating the UE selected for transmission in each TTI. Note that we reuse the same policy for different RBGs to avoid an exponential increase of action space, which may incur significant training costs and probably bad convergency.

\textbf{Reward}: The long-term KPIs, i.e., THP, JFI and PDR as calculated in Eq. (\ref{eq:kpi}), are the final performance concerns and can be calculated only at the end of each scheduling duration. We need to consider them in each scheduling step by defining a proper reward function. A straightforward definition is in the form of linear weighted sum, i.e.,
\begin{equation}
\label{eq:reward}
r = \alpha \cdot \mathrm{thp} + \beta \cdot \mathrm{jfi} - \delta \cdot \mathrm{pdr}
\end{equation}
where $\mathrm{thp}$, $\mathrm{jfi}$ and $\mathrm{pdr}$, which are obtained at each TTI after the scheduling decision action is executed, can be viewed as a single-step version of THP, JFI and PDR. $\mathrm{thp}$ is the total throughput of the current TTI. $\mathrm{jfi}$ is the Jain's fairness index calculated from the beginning of the scheduling duration to the current TTI. $\mathrm{pdr}$ is the total number of dropped packets in the current TTI, and the value is normalized with $K$. $\alpha$, $\beta$ and $\delta$ are the weighting factors. Although this linear scalarization of a multi-objective problem may lead to non-convex Pareto frontier, we find it still feasible to obtain satisfactory results at least in this paper.

\subsection{Scheduling Scheme}
As the scheduling problem has a continuous state space as defined above, value-based DRL methods, such as Deep Q-Network (DQN), are inefficient. To handle this issue, we employ the advantage actor-critic (A2C) framework which is essentially a policy-based DRL algorithm. The policy $\pi_\theta(a|s)$ is directly optimized by adjusting the parameters $\theta$. Similar to the actor-critic algorithm, the A2C algorithm employs two NNs, i.e., one policy network to make the decision and one value network for judging the decision.

At the $t$-th TTI of the scheduling duration, the DRL agent observes state $s_t$ from the environment, and makes scheduling decision $a_t$. After the system executes the decision, the DRL agent receives the reward $r_t$ and observes the next state $s_{t+1}$. The goal of DRL training is to find a policy $\pi\left( a|s\right)$ through interactions with the environment to maximize the accumulated (discounted) reward $R$
\begin{equation}
\label{eq:expreward}
R= \sum_{t=0}^{\infty} \gamma ^ t r_t
\end{equation}
where $\gamma$ is the discount factor to determine the importance of the future reward.

The parameters $\theta$ can be updated by gradient ascent on the expected return $R$
\begin{equation}
\label{eq:gradient}
g=\nabla_\theta \mathbb{E}\left[R\right] = \nabla_\theta \mathbb{E}\left[\sum_{t=0}^{\infty} \gamma ^ t r_t \right]
\end{equation}
The gradient in \eqref{eq:gradient} can be further represented as \cite{schulman2015high}:
\begin{equation}
\label{eq:gae}
g = \mathbb{E}\left[ \sum_{t=0}^{\infty}A^\pi \left(s_t, a_t\right) \nabla_\theta \log \pi_\theta \left(a_t | s_t\right)  \right]
\end{equation}
where, $A^\pi \left(s_t, a_t\right)$ is the advantage function. Instead of using one-step temporal difference (TD) as \cite{schulman2015high}, we introduce the $n$-step TD for the advantage function
\begin{equation}
\label{eq:nstep}
\begin{split}
A^\pi \left(s_t, a_t\right) = \mathbb{E}_{s_{t+1}\cdots s_{t+n}} [&r_t + \gamma r_{t+1} + \gamma^2 r_{t+2} +\cdots \\
&+ \gamma ^ n V^\pi\left(s_{t+n}\right)- V^\pi\left(s_t\right)]
\end{split}
\end{equation}
where more than one steps are considered to get $n$ rewards along the trajectory. In the scheduling problem with long-term KPIs, the $n$-step TD calculation helps to improve the advantage estimation by averaging out the variance during gradient updates and leads to a more stable training. Through simulation, it is found that setting $n=20$ is good enough.

In order to avoid the agent from trapping in a deterministic local optima, the entropy regularization in \eqref{eq:entropy} is employed to enhance the exploring ability
\begin{equation}
\label{eq:entropy}
H=-\sum\limits_a \pi_\theta(a|s)\log\pi_\theta(a|s)
\end{equation}

To improve the generalization capability, the DRL agent is trained on several environments with different deployments simultaneously.

For multi-RBG cases, the policy network can be reused iteratively for each RBG to deal with the dimension curse. The only thing needs to be considered is that the states for scheduling different RBGs in the same TTIs should be updated according to the scheduling decisions made for the previous RBGs.

To sum up, the full algorithm is described in Algorithm ~\ref{alg:a2c}. A batch of training data is comprised of $n$ consecutive interactions with the environment. The advantage function of each experience is then obtained using \eqref{eq:nstep}. $\theta$ and $\phi$ are the parameters of the policy network and the value network. Finally, parameters are updated through the gradient decent (GD) method.

\begin{algorithm}[!htbp]
	\caption{A2C algorithm}
	\label{alg:a2c}
	\begin{algorithmic}
		\State Initialize all environments
		\State Initialize policy network $\pi_\theta$ and value network $V_\phi$
		\State Initialize experience buffer $E$
        \State Set maximum number of training episode $W=10000$, entropy weight $\lambda_e=0.03$, value weight $\lambda_v=0.5$
		\For {iteration = 1 to $W$}
		\State \Call{Sample\_Batch}{$n, \pi_\theta$}
		\State Update advantage $A_i$ for $i$th experience in $E$
		\State Policy objective $J_\theta = \sum_{i} A_i \log \pi_\theta (a_i | s_i) $
		\State Entropy term $H_\theta = - \sum_{i}\pi_\theta(a_i|s_i) \log \pi_\theta(a_i|s_i)$
		\State Loss of value $L_\phi = \sum_{i} A_i ^2$
		\State GD with $G = - \left(\nabla_\theta J_\theta + \lambda_e \nabla_\theta H_\theta \right) + \lambda_v\nabla_\phi L_\phi$
		\EndFor
		
		\Function{Sample\_Batch}{$n, \pi_\theta$}
		\State Clear $E$
		\For {t = 1 to $n$}
		\State Observe $s_t$
		\State Choose action $a_t \sim \pi_\theta(s_t)$
		\State Take action $a_t$, observe $s_{t+1}$ and $r_t$
		\State Store ($s_t, a_t, r_t, s_{t+1}$) into $E$
		\EndFor
		\EndFunction
	\end{algorithmic}
\end{algorithm}

\subsection{Neural Network Design}
The most straightforward NN design for the DRL agent is using fully connected networks with states of all UEs together serving as the input. We call it one-pass design to distinguish from the scalable design elaborated later. For the one-pass NN design, rectified linear unit (ReLU) function is used as the activation function for all hidden layers in both the policy and the value network. For the activation function of the output layer, softmax is used for the policy network and linear function is used for the value network.

\begin{figure}[!t]
	\centering
	\includegraphics[width=8cm]{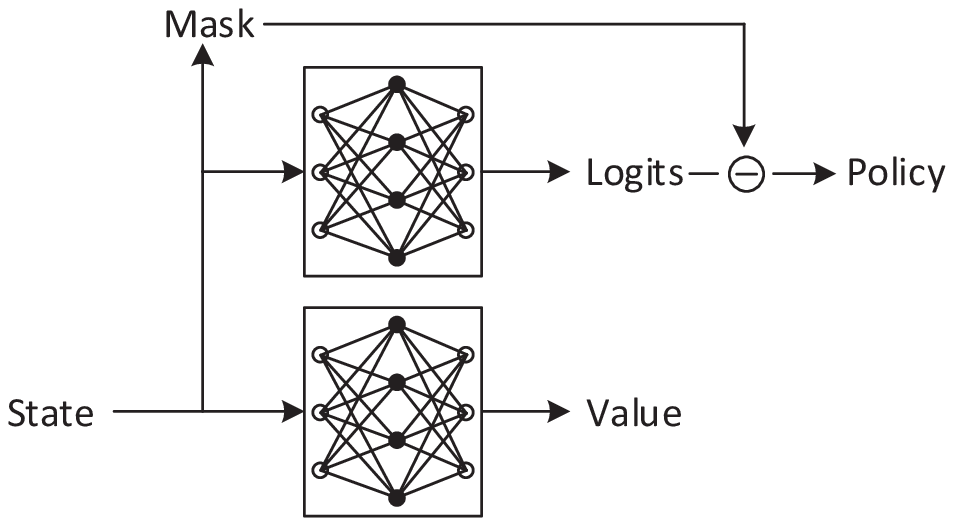}
	\caption{One-pass NN design.}
	\label{fig:onepass}
\end{figure}

As introduced in Section \ref{sec3}, for non-full buffer traffic mode cases, a UE may be inactive when its RLC buffer is empty and become active again when new packets arrive. To handle this situation, we introduce a mask processing in the one-pass NN design as shown in Fig. \ref{fig:onepass}. A mask is generated by the states, and prohibits policy from choosing the inactive UE(s), e.g., by subtracting a large value from the corresponding logits.

In the one-pass NN design, states of all UEs are input together into the NNs, which makes the scale of NNs dependent on the number of UEs. For instance, the input layer dimensions of both the policy and the value network are $4K$ with $K$ UEs and 4 state parameters for each UE. When adopted in different deployments, the number of UEs may be different and changing now and then. Setting a maximum possible number of UEs $K_{\max}$ may be a solution, but it may result in unnecessary high computational complexity when $K < K_{\max}$ and failure to work when $K > K_{\max}$.

\begin{figure}[!t]
	\centering
	\includegraphics[width=9cm]{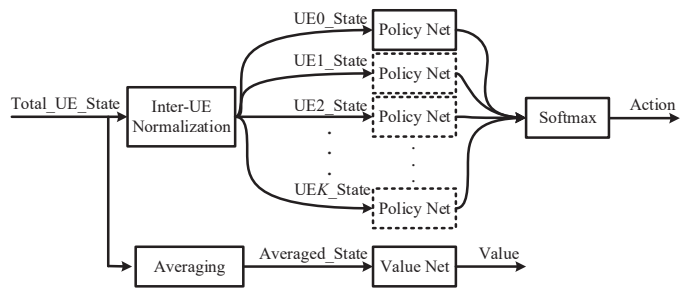}
	\caption{Scalable NN design.}
	\label{fig:scalable}
\end{figure}

We propose a scalable NN design, where the same policy network is repeated for $K$ times with handling states of one UE each time, and a value network is used only once to handle the average states of all UEs. The same policy network is shared among multiple UEs. The details of the scalable NN design is shown in Fig. \ref{fig:scalable}, where both the policy and the value network are fully connected ones and the activation functions are the same to the one-pass design. Obviously, the input layer dimensions of both the policy and the value network are 4, independent of the number of UEs. Similarly, the output layer dimensions of both networks are 1. When the number of UEs varies, the only impact is the number of usage of the policy network. The theoretical explanation behind the scalable NN design is the so called permutation invarianace property of the scheduling task. According to \cite{Zaheer2017deep}, if any permutation of the inputs of a function leads to the same permutation of its outputs, the function is said to be invariant to permutation. For the considered scheduling task, if the states of multiple UEs are permuted, the scheduling metrics will also be permuted accordingly. To exploit this permutation invariance property, the same NN can be shared among all the UEs just as the proposed scalable NN design does.

\subsection{Virtual Environment Training}
\begin{figure}[!t]
	\centering
	\includegraphics[width=9cm]{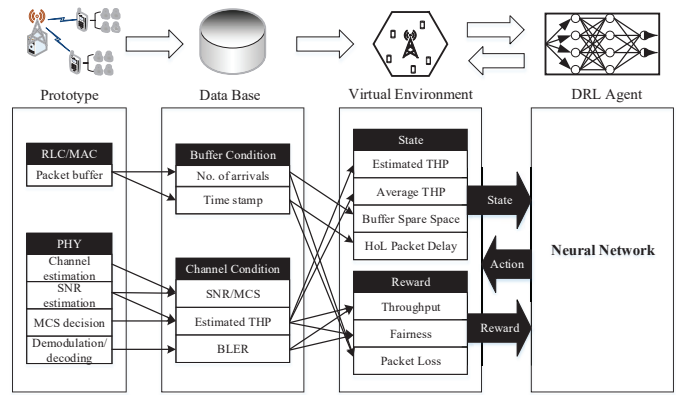}
	\caption{Virtual environment training framework.}
	\label{fig:ve}
\end{figure}

Different from the application scenarios like Atari Games, the performance and robustness requirements of communication systems are far more higher. The training of DRL agents usually costs a long time, during which the performance and robustness may be degraded due to the exploration of new actions. Directly adopting a randomly initialized agent to the practical system and training it from scratch on line are quite inefficient. Instead, we propose to train the agent in a virtual environment first and set the trained NN parameters as the initial version of the agent adopted in the practical system.

The virtual environment training framework is shown in Fig. \ref{fig:ve}. The data, including channel and buffer conditions, are collected from the practical systems and stored in a data base. The virtual environment is then established based on these collected data, and provides the necessary informations to the DRL agent for training. The virtual environment can generate training samples much faster than the practical system, helping with the fast convergence of the DRL agent.

\section{Evaluations and Discussions}
\label{sec6}
The proposed DRL scheduling scheme is evaluated through both simulation and field test.
\subsection{Simulations}
\label{sec6.1}
\begin{figure}[t]
	\centering
	\includegraphics[width=9cm]{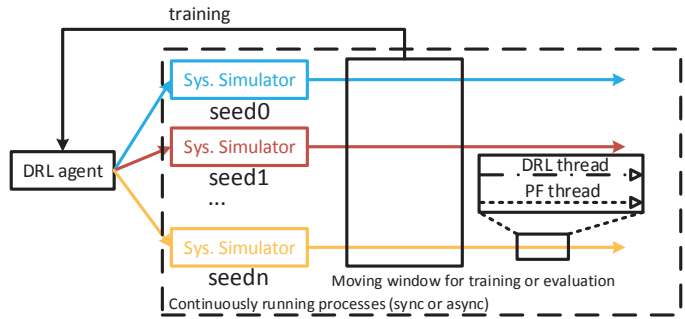}
	\caption{Simulation framework.}
	\label{fig:framework}
\end{figure}

As shown in Fig.~\ref{fig:framework}, the simulation platform contains a DRL agent and several system simulators working in parallel as the environment. The system simulator is calibrated according to the physical layer (PHY) and medium access control layer (MAC) standard of LTE, where AMC and OLLA are all implemented. Each system simulator provides a training environment with a specific UE deployment. Two threads are initialized for each simulator: one interacts with the DRL agent, and the other executes PF scheduling algorithm as baseline for performance comparison. A moving window is used to restrict the time scope of training and evaluation. The DNNs are updated at the end of every scheduling duration, and performance comparison is carried out every 50 updates. The simulation settings are listed in Table \ref{tab:simulationsetting}.

\begin{table}[h]
	\centering
	\caption{Simulation Settings}
	\begin{tabular}{c|c}
		\hline
		Parameters		    &   Values \\
		\hline
		Carrier Frequency	&	2GHz \\
        \hline
        Channel Model       &   SCM-3D-UMA \\
        \hline
        TTI duration        &   1ms \\
        \hline
        Scheduling Duration &   500TTIs \\
        \hline
        Bandwidth           &   1.4MHz(single RBG), 20MHz(multiple RBGs) \\
        \hline
        Number of RB        &   6(single RBG), 100(multiple RBGs) \\
        \hline
        Number of RBG       &   1, 10 \\
        \hline
        Number of UE        &   80\% indoor, 20\% outdoor \\
        \hline
        UE speed            &   3km/h(indoor), 30km/h(outdoor) \\
        \hline
        Antenna Setting     &   4T2R with 1 stream \\
        \hline
        OLLA type           &   fixed-step OLLA (0.1dB step) \\
        \hline
        Packet Arrival Rate &   200 per second for all UEs \\
        \hline
        Buffer Size         &   1e6 bits \\
        \hline
        Buffer type         &   FIFO \\
        \hline
        Packet Size         &   8e3 bits \\
        \hline
        Maximum Delay       &   2s \\
		\hline
	\end{tabular}
	\label{tab:simulationsetting}
\end{table}

The fully connected NNs used for the DRL agent contains two hidden layers, each of which has $128V$ neurons with $V=K$ for the one-pass NN design and $V=1$ for the scalable NN design. The learning rate is set to 0.001 at the beginning and decays by the rate 0.1 after 5000 updates. The discount factor in Eq. \ref{eq:expreward} is set to 0.9. The weights used in the reward function, i.e., Eq. \ref{eq:reward}, are $\alpha=0.07$, $\beta = 0.71$, $\delta = 0.22$.

\subsubsection{Simulation evaluation on the one-pass NN design}
We first evaluate the performance of one-pass NN design. Since it can only handle scenario with fixed-number of UEs, we consider 5 UEs in the simulation. 56 simulators for a same DRL agent are launched simultaneously for a quickly and adequately exploring in state and action space. The deployment of the UEs and random seeds for the simulators are deliberately differentiated to undermine the data correlation that would harm the DRL. In addition, averaging among different random seeded simulators also increases the generalization of the DRL agent and are more reasonable in the sense of performance comparison.

%

\begin{figure}[!t]
	\centering
	\includegraphics[width=8cm]{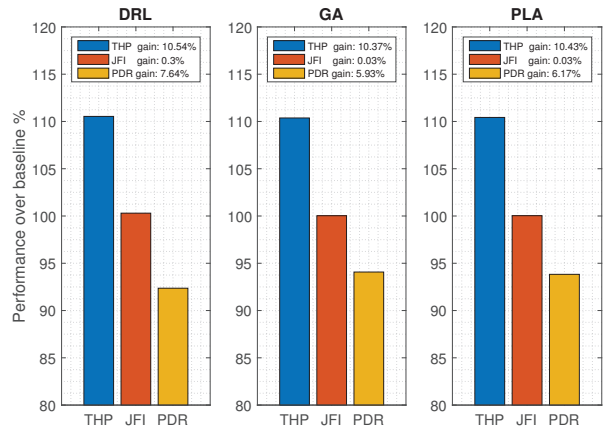}
	\caption{Performance of DRL with one-pass NN design and genie-aided methods for single RBG.}
	\label{fig:singlerbg}
\end{figure}

\begin{figure}[!t]
	\centering
	\includegraphics[width=8cm]{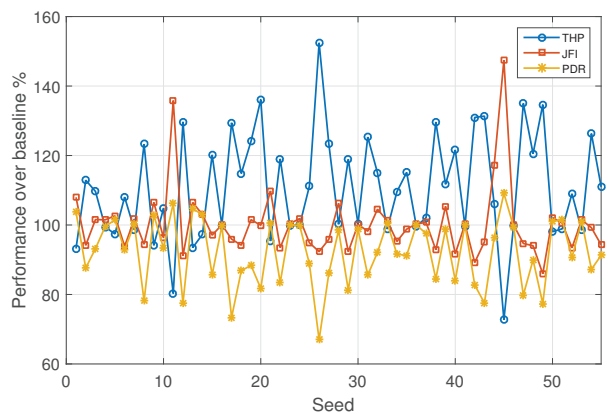}
	\caption{Performance metrics for DRL with one-pass NN design of each seed.}
	\label{fig:seed}
\end{figure}

\begin{figure}[!t]
	\centering
	\includegraphics[width=8cm]{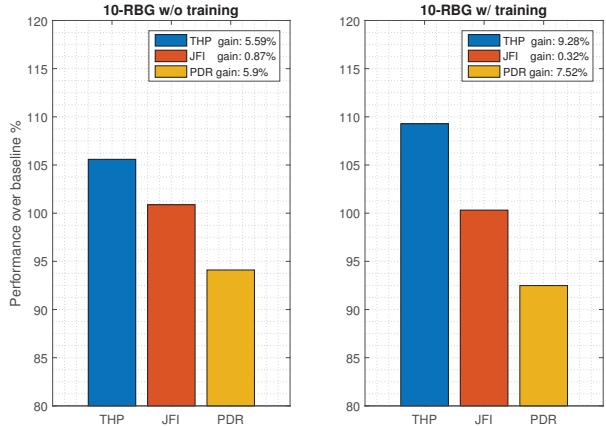}
	\caption{Performance of DRL with one-pass NN design for multiple RBGs.}
	\label{fig:multirbg}
\end{figure}

After $\sim$5000 updates, we fix the parameters of the NN model and run a performance evaluation. The average performances over baseline in 20000 TTIs of 56 UE deployments are shown in Fig.~\ref{fig:singlerbg} and Fig.~\ref{fig:multirbg}, where the PF algorithm is used as the baseline. It is worth noting that values above 100\% for THP and JFI, and below 100\% for PDR mean better performance, i.e., higher throughput, better fairness and lower packet loss.

For single RBG case, we also provide the results of GA and PLA, in which the states from the simulators are recorded in advance and fed into the algorithms. For GA, we use simulated binary crossover operator and polynomial mutation with the crossover probability of $p_c = 0.95$ and a mutation probability of $p_m = 0.05$. The distribution indexed for crossover and mutation operators are $\eta_c = 5$ and $\eta_m = 20$, respectively. We set the population size as $M=3000$ and $G=10000$ generations to conduct the GA. For PLA, the maximum list number of PLA is $L_{\max}=2000$. From Fig.~\ref{fig:singlerbg}, we can see that the performances of GA and PLA are similar. The DRL scheme obtains nearly the same throughput, and slightly better JFI and PDR performance compared to the two gene-aided algorithms. It is worth noting again that the two genie-aided methods are included herein to show the possible gain space over the baseline. They are impractical due to the assumption about knowing future channel conditions and buffer states. It is quite encouraging to see that the proposed DRL scheme learns only one policy for all UE deployments and achieves similar gain over baseline without knowing future information.

Fig.~\ref{fig:seed} plots the performance of DRL in all 56 UE deployments, where we can see that the THP and PDR gain are obvious among all deployments while JFI keeps almost the same to the baseline. We argue that some THP vales, e.g., seed 11 and seed 45, are not failed because they have larger JFI, meaning that they are still somewhere near the Pareto frontier. We believe that in real world, some deployment-specific KPI weightings will help with fast converge to the required performance.

For multiple RBG scheduling, two methods have been tried: a) transfer learning, i.e., the NNs trained for single RBG case is directly reused. b) retraining the model that fits for multi-RBG case. From Fig.~\ref{fig:multirbg}, it is found that the first method still works. This justifies that the good generalization capability of the trained model. The retraining method further exploits the learning ability of the agent and achieves a better performance similar to the single RBG case.

\subsubsection{Simulation evaluation on the scalable NN design}
\begin{figure}[!t]
	\centering
	\includegraphics[width=8cm]{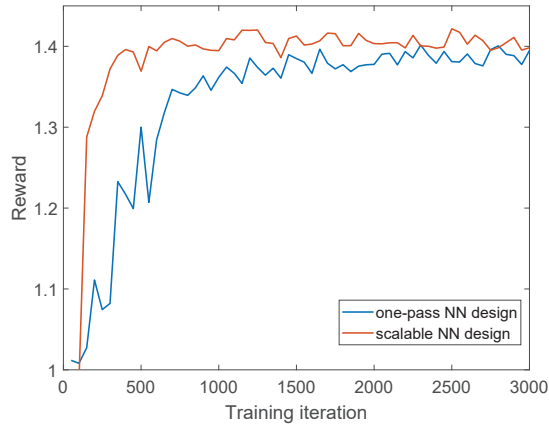}
	\caption{Convergence of scalable NN design.}
	\label{fig:converge}
\end{figure}

We then evaluate the performance of the scalable NN design. Again, deployments with 5 UEs are used for the agent training. Fig. \ref{fig:converge} shows the converging speed comparison between the one-pass NN design and scalable NN design. Although the two designs achieve similar rewards in the end, the convergence for the scalable design is much faster. It is because that the policy network is used for $K$ times in each TTI, which is equivalent to have $K$ training samples, while only one training sample is produced for the one-pass design in the same time interval.

\begin{figure}[!t]
	\centering
	\includegraphics[width=8cm]{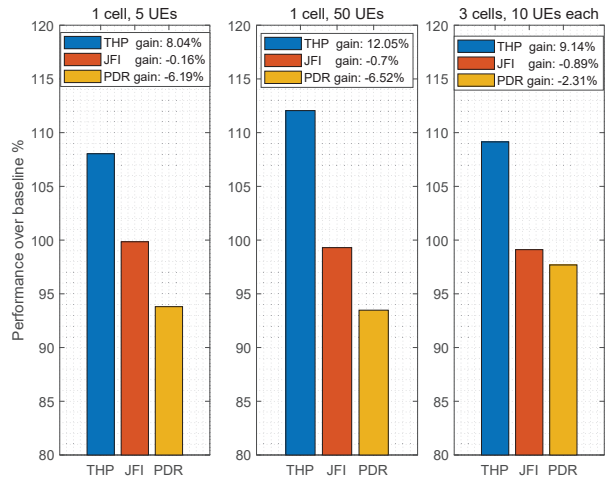}
	\caption{Performance of DRL with scalable NN design.}
	\label{fig:scaleperf}
\end{figure}

The performance of the scalable NN design is shown in Fig. \ref{fig:scaleperf}. It is shown that its performance is similar to the one-pass design in the 5-UE scenario. Then, the DRL agent with the scalable NN design trained in the 5-UE scenario is directly adopted in a 50-UE single cell scenario and 3-cell scenario with 10 UEs per cell. The performance gains over the baseline can also be obtained in the latter two scenarios. The simulation results verified that the scalable design could be adopted in scenarios with different UE numbers and deployments.

\subsection{Field Tests}
\label{sec6.2}
The ultimate goal of design the smart scheduling scheme is to use it in the practical system. Hence, in this section, we implement the scheme into a field test prototype system and conduct field trials. Due to its benefits over the one-pass NN design verified by the simulations, we only test the scalable NN design in the field trials.

\begin{figure}[!t]
	\centering
	\includegraphics[width=8cm]{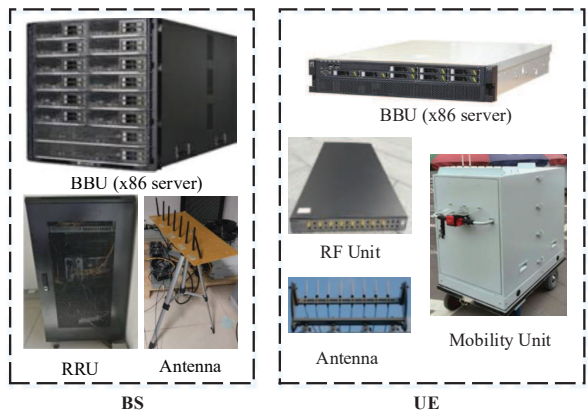}
	\caption{Field test prototype.}
	\label{fig:prototype}
\end{figure}

The prototype system includes one BS and 8 UEs. The baseband processing functions of the BS and UEs are implemented in x86 servers, as shown in Fig. \ref{fig:prototype}, for the sake of easy configuration. The UEs' positions are fixed, and the average signal-to-noise ratios (SNRs) for the UEs are shown in Table \ref{tab:uesnr}.
\begin{table}[h]
	\centering
	\caption{Average SNRs for the UEs}
	\begin{tabular}{c|c|c|c|c|c|c|c|c}
		\hline
		UE ID		& 0     & 1     & 2     & 3     & 4    & 5    & 6    & 7 \\
		\hline
		SNR (dB)	& 17.6 & 18.2 & 17.9 & -0.7 & 9.8 & 9.7 & 8.7 & -2.3\\
		\hline
	\end{tabular}
	\label{tab:uesnr}
\end{table}

Other field test settings are listed in Table \ref{tab:fieldtestsetting}, where only 80 RBs are used for data transmission with other RBs on the upper and lower sideband are left empty. The scalable OLLA adjusts the modulation and coding scheme (MCS) directly according to the ACK/NACK feedback.
\begin{table}[h]
	\centering
	\caption{Field Test Settings}
	\begin{tabular}{c|c}
		\hline
		Parameters		    &   Values \\
		\hline
		Carrier Frequency	&	3.59GHz \\
        \hline
        Scheduling period   &   1 frame (10ms) \\
        \hline
        Bandwidth           &   20MHz \\
        \hline
        Number of RB        &   80  \\
        \hline
        Number of RBG       &   1, 5 \\
        \hline
        Subcarrier Spacing  &   15kHz \\
        \hline
        Number of UE        &   8 \\
        \hline
        Antenna Setting     &   1T1R \\
        \hline
        OLLA type           &   Scalable OLLA \\
        \hline
        Packet Arrival Rate &   200 per second for all UEs \\
        \hline
        Buffer Size         &   2e3 packets \\
        \hline
        Buffer type         &   FIFO \\
        \hline
        Packet Size         &   3e4 bits \\
        \hline
        Maximum Delay       &   2s \\
		\hline
	\end{tabular}
	\label{tab:fieldtestsetting}
\end{table}

\begin{figure}[!t]
	\centering
	\includegraphics[width=8cm]{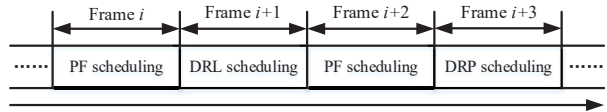}
	\caption{Frame structure used in field tests.}
	\label{fig:frame}
\end{figure}

To do a fair comparison between the proposed smart scheduling and PF scheduling, we should ensure that the two schemes experience exactly the same channel and RLC buffer conditions. For the channel conditions, we implement a novel frame structure as shown in Fig. \ref{fig:frame}, where the two schemes are carried out alternatively. With channel changing slowly, the channel conditions experienced by the two schemes can be viewed as the same. For the RLC buffer conditions, we maintain two set of RLC buffers, each for one scheme. The packet arrivals for the two sets of RLC buffers are exactly the same. In the frames with smart scheduling scheme, the corresponding set of RLC buffers provides buffer related states, whereas when it comes to the frames with PF scheduling, the other set of RLC buffers is adopted.

We consider two test scenarios, one with the number of UEs fixed to 8, and the other with a variable number of UEs. In each test scenarios, both single-RBG and multiple-RBG cases are designed. Each test lasts for about 15 minutes. For the scenario with variable number of UEs, 8 UEs participate in the scheduling at the beginning. After about 5 minutes, UE 1, 2, 3 and 6 are turned off, then after another 5 minutes, they are turned on again. The test cases can be summarized as Table \ref{tab:testcase}.

\begin{table}[h]
	\centering
	\caption{Test Case Design}
	\begin{tabular}{c|c|c}
		\hline
		Case ID	 &  UE number & RBG number \\
		\hline
		Case 1	 &	  8       &    1       \\
        \hline
        Case 2   &    8       &    5       \\
        \hline
        Case 3   &   variable &    1       \\
        \hline
        Case 4   &   variable &    5       \\
		\hline
	\end{tabular}
	\label{tab:testcase}
\end{table}

\begin{figure}[!t]
\centering
\includegraphics[width=8cm]{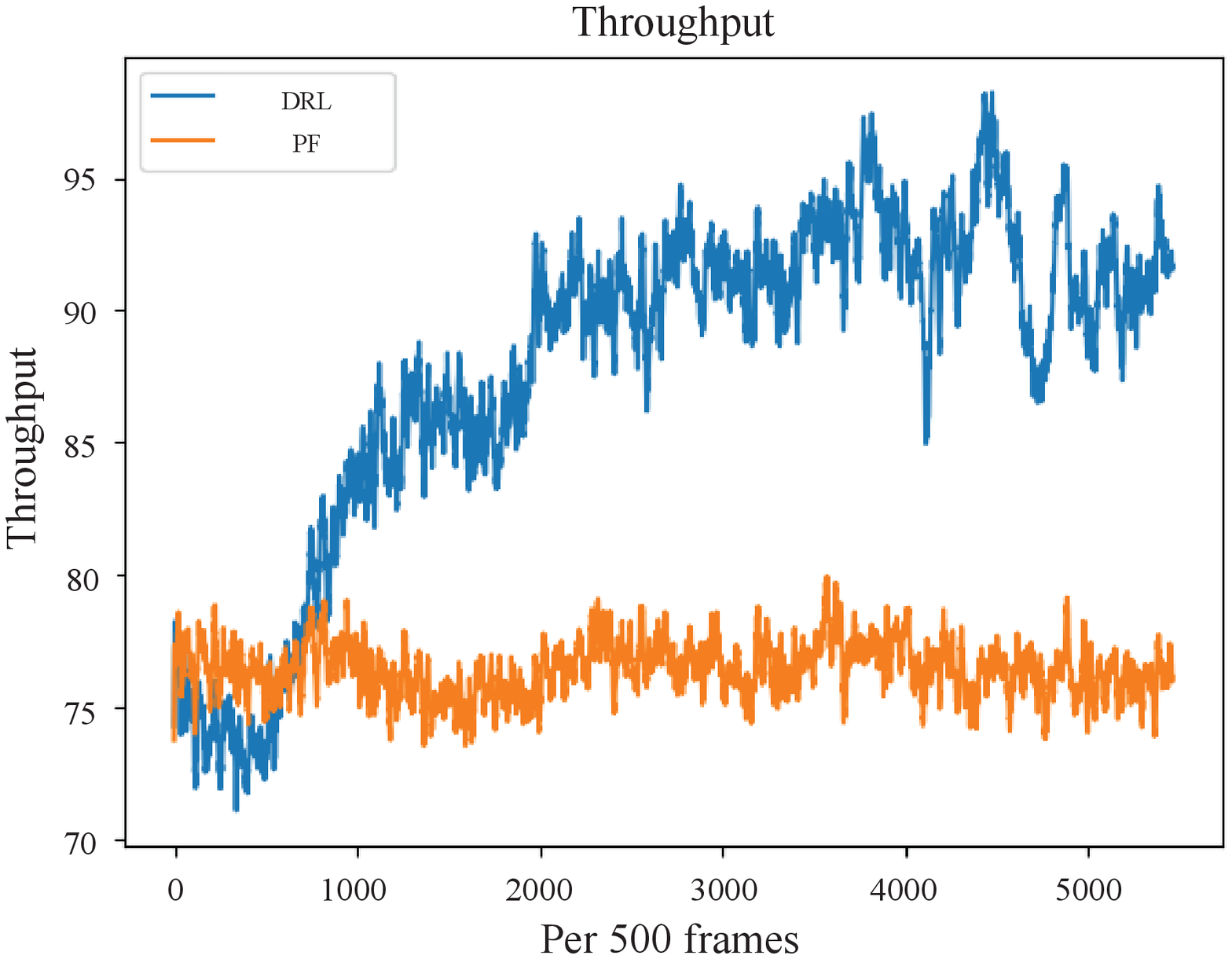}%
\hspace{2in}%
\includegraphics[width=8cm]{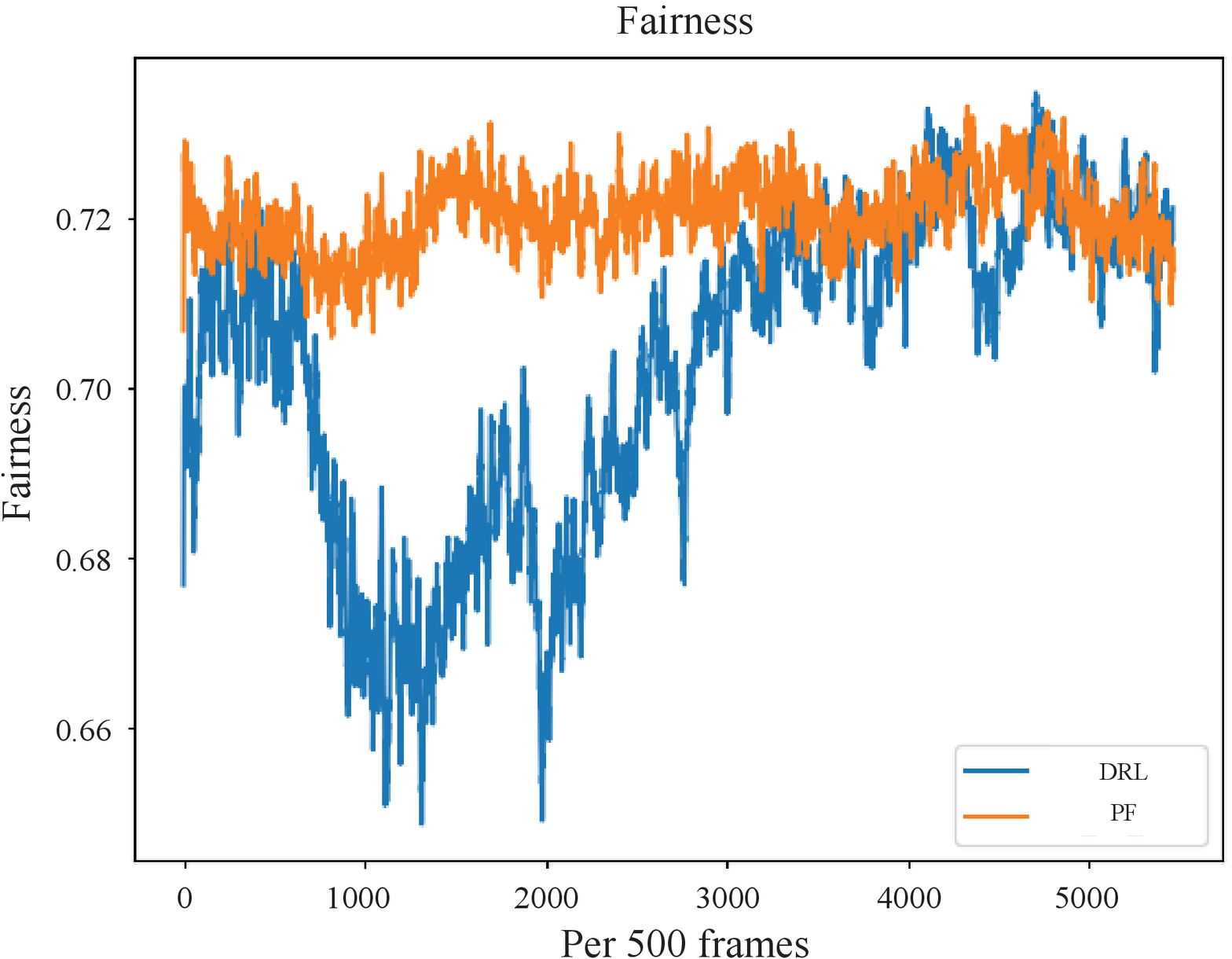}
\caption{Performance during training DRL agent from scratch.}
\label{fig:trainOnLine}
\end{figure}

First of all, we try to train the DRL agent online from scratch with 8 UEs. Since the agent in the field trial is facing only one deployment, and the timing must obey the operations of the practical system, the gathering of training samples is quite low-efficiency. We draw the throughput and fairness performance during training in Fig. \ref{fig:trainOnLine}, where each point of the x-axis represents an averaging result over 10 seconds, i.e., 500 frames for the smart scheduling and 500 frames for PF scheduling. From the result, it can be observed that although the throughput performance of the smart scheduling scheme quickly outperforms the baseline, 35000 seconds, i.e., about 10 hours, are needed for the agent to find a policy with comparable fairness performance compared with the baseline. It is obviously unacceptable and unrealistic to spend such a long time to do the training. Hence, we rely on the virtual environment training framework for the following field tests. The DRL agent is trained offline in the virtual environment first, and then used in the practical testing prototype.

Then, we conduct the field trials and the test results are shown in Table \ref{tab:testresult}, where the number of dropped packets are recorded instead of the PDR and the minus gain means dropping less packets from the RLC buffer. Generally, with similar JFI performance, the proposed smart scheduling has higher throughput and less dropped packets in all test cases. In one RBG cases, the throughput gain of the smart scheduling scheme is more than 20\%, while for the cases with 5 RBGs, the gains shrink to about 9\%. It is because with more RBG divided, the channel measurement is more precise for each RBG, which helps both PF and smart scheduling scheme with achieving high throughput, hence, the gain space left becomes less. We do not see much difference for the smart scheduling scheme between fixed UE number cases and variable UE number cases, which verified the feasibility of the scalable NN design.
\begin{table}[!htbp]
	\centering
	\caption{Field Test Results}
	\begin{tabular}{c|c|c|c|c}
		\hline
		Case ID	 &  KPI       &  \multicolumn{2}{|c|}{Scheme} & Gain    \\ \cline{3-4}
                 &            &  PF      &  DRL               &         \\
		\hline
		    	 &	THP(Mbps) &  75.67   &  93.61             & 23.7\%  \\ \cline{2-5}
        Case 1   &  JFI       &  0.719   &  0.730             & 1.5\%   \\ \cline{2-5}
                 &  Dropped packet number      &  657.87   &    571.62         &  -13.1\%  \\
        \hline
		    	 &	THP(Mbps) &  96.42   &  105.51            & 9.4\%   \\ \cline{2-5}
        Case 2   &  JFI       &  0.746   &  0.755             & 1.2\%   \\ \cline{2-5}
                 &  Dropped packet number       &  594.32   &    552.81          &  -7.0\% \\
        \hline
		    	 &	THP(Mbps) &  74.67   &  91.89             &  23.1\%  \\ \cline{2-5}
        Case 3   &  JFI       &  0.714   &  0.730             &  2.3\%   \\ \cline{2-5}
                 &  Dropped packet number       &  578.32  &     466.74         &  -19.3\% \\
        \hline
		    	 &	THP(Mbps) &  92.99   &  101.99            &  9.7\%   \\ \cline{2-5}
        Case 4   &  JFI       &  0.758   &  0.747             &  -1.5\%  \\ \cline{2-5}
                 &  Dropped packet number       &  508.98  &    415.75        &  -18.3\%   \\
		\hline
	\end{tabular}
	\label{tab:testresult}
\end{table}

\section{Conclusions and Future Work}
\label{sec7}
In this paper, a smart scheduling scheme based on DRL is proposed for cellular networks. Multiple long term optimization objectives, i.e., THP, JFI and PDR, are considered in the problem. Through both the simulation and the field test evaluation, we have the following observations
\begin{enumerate}
  \item The smart scheduling scheme using A2C algorithm can achieve higher throughput and lower packet loss with nearly no degradation of fairness compared with the PF scheduling in the non-full buffer traffic mode. Its performance is similar to the two gene-aided algorithms, while no future state information is needed.
  \item With the help of the scalable NN design, the DRL agent has both fast convergence and good generalization capability. It can track the dynamics of the cellular networks and be adopted in practical scenarios with different system scales.
  \item Instead of training the DRL agent from scratch directly in the practical system, it is preferred to do the training in a virtual environment first. The pre-trained agent can be used as the initial version for the practical system, and prevent the performance and robustness degradation.
\end{enumerate}

Multi-cell scenarios are tried in our simulation, however inter-cell interference coordination cannot be performed through the proposed single agent design. In the future, multi-agent reinforcement (MARL) can be tried and even more performance gain may be expected.

{\footnotesize
\bibliographystyle{IEEEtran}
\bibliography{drlschedulingjournal}
}

\end{document}